\RequirePackage{fix-cm}
\documentclass{svjour3}                     

\smartqed  
\usepackage{graphicx}
\usepackage{amsmath}
\usepackage{float}
\usepackage[caption = false]{subfig}
\usepackage{graphicx}
\usepackage[title]{appendix}
\usepackage{natbib}
\usepackage{epstopdf}
\usepackage{algorithm}
\journalname{ArXiV submission}
\begin{document}

\title{A Staged Approach to Evolving Real-world UAV Controllers}


\author{Gerard David Howard         \and
        Alberto Elfes 
}


\institute{G.D. Howard \at
              QCAT, 1 Technology Court Pullenvale 4069 QLD Australia \\
              Tel.: +61-7-3327-4714\\
              ORCID: 0000-0002-5012-7224\\
              \email{david.howard@csiro.au}           
           \and
           A. Elfes \at
              QCAT, 1 Technology Court Pullenvale 4069 QLD Australia \\
              ORCID: 0000-0003-2433-995X\\
}

\date{26th May 2019}

\maketitle

\begin{abstract}
A testbed has recently been introduced that evolves controllers for arbitrary hover-capable UAVs, with evaluations occurring directly on the robot.  To prepare the testbed for real-world deployment, we investigate the effects of state-space limitations brought about by physical tethering (which prevents damage to the UAV during stochastic tuning), on the generality of the evolved controllers.  We identify generalisation issues in some controllers, and propose an improved method that comprises two stages: in the first stage, controllers are evolved as normal using standard tethers, but experiments are terminated when the population displays basic flight competency.  Optimisation then continues on a much less restrictive tether, effectively free-flying, and is allowed to explore a larger state-space envelope.  We compare the two methods on a hover task using a real UAV, and show that more general solutions are generated in fewer generations using the two-stage approach.  A secondary experiment undertakes a sensitivity analysis of the evolved controllers.
\end{abstract}


\section{Introduction}
\label{sec_intro}

Rapidly-advancing Additive Manufacturing technologies are leading a shift from mass production of identical simulacrums to provision of one-off, bespoke systems.  Robotics is particularly suited to benefit from this shift; permitting rapid prototype-and-test of bespoke robotic systems with specialised morphologies to achieve heightened environmental/task-specific performance.  In such an approach, robots may appear in a wide variety of different morphological compositions, with varied payloads.  Behaviours must be generated to fully harness this diversity of morphologies.  Evolutionary Algorithms (EAs) are ideal for this task as they are problem- and platform-agnostic optimisers that can account for morphology, payload, environmental and task requirements~\citep{eiben_smith}.  

Although EAs are a promising suite of optimisers for this task, their dependence on stochastic search processes leads to difficulties when optimising on real hardware, mainly around time requirements, repeatability, and accurate fitness assessment.  As a consequence, the push towards bespoke robotics requires a concurrent push towards automated learning facilities (herein referred to as {\em testbeds}~\citep{heijnen2017icra,auto_increm_evo}), where robots will have their behaviours and morphologies optimised.  Key characteristics of such testbeds include: closed loop operation (no human intervention required), robot-ambivalent optimisation, and the ability to produce robots that work in the real world.

UAVs are an ubiquitous example of general-purpose robots being required to perform a rapidly-expanding variety of increasingly specific tasks (e.g., tracking, mapping, inspection, repair, and delivery).  As these tasks become more challenging (or more niche), the ability of a general-purpose UAV to perform well significantly decreases.  As such, UAVs are highly likely to benefit from the development of systems that permit their specialisation for increased task performance, an increased range of deployment scenarios, and more favourable mission outcomes.  

\begin{figure*}[ht]
\centering
\subfloat[]{\includegraphics[width = 6.5cm]{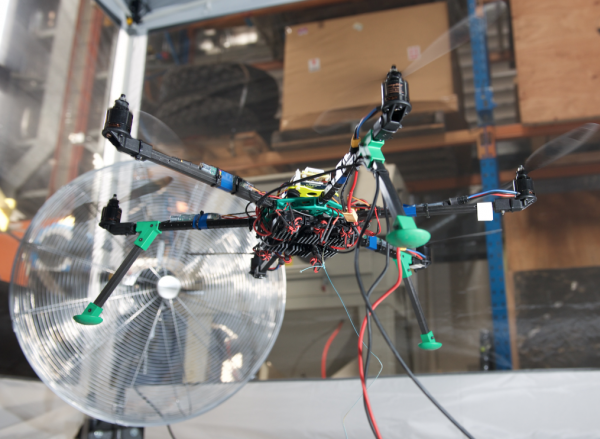}}
\subfloat[]{\includegraphics[width = 6.5cm]{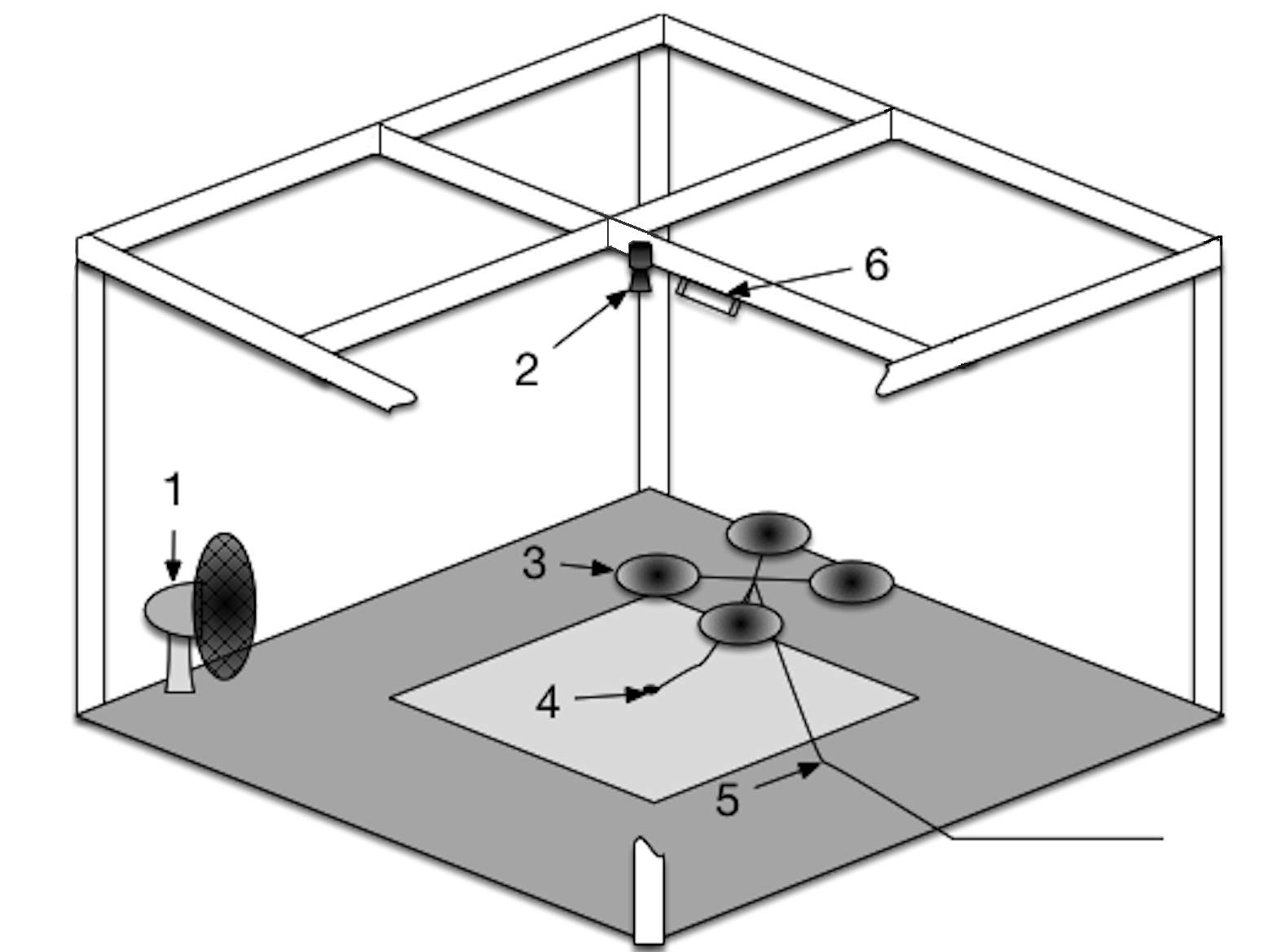}}
\caption{(a) Showing a hexacopter in flight. (b) The testbed, showing (1) the fan, (2) camera, (3) UAV, (4) physical tether, (5) data/power tether, and (6) light.  The camera height is 200cm and padded floor area is 271cm$^2$.}
\label{setup_new}
\end{figure*}

To address this need, we recently introduced a testbed (Fig.~\ref{setup_new}) that optimises controllers for arbitrary UAVs~\citep{howard2015platform,howard2017platform}\footnote{Herein we use 'UAV' to refer to any hover-capable multirotor, with an airframe size $<800mm$}.   To date, we have already demonstrated (i) repeatable evolutionary experimentation to test hypotheses in the real world, and (ii) optimizing UAVs with unconventional payloads and physical setups that would not be otherwise flyable.  

This article covers the development of our testbed from a prototyping to a deployment stage, ensuring that {\em our testbed reliably generates controllers that are more generalised to real-world conditions}.  The main scientific contributions of this work are:

\begin{enumerate}
\item The development of a two-staged technique (a form of incremental learning~\citep{Rossi2014}) that works in hardware on real UAVs and reliably generates controllers suitable for real UAV deployments;
\item Statistical comparisons between the new and old methods, and;
\item A comprehensive analysis of the best controllers to ascertain how close they are to optimal values.
\end{enumerate}

We pose the following research questions:

\begin{enumerate}
\item Given that evolutionary performance is critically dependent on the amount of state space available~\citep{SaB}, can we improve the optimisation process by increasing the amount of state space experienced during training enough that the evolved controllers can generalise to unseen waypoints, as found in real world missions?  Can we balance this desire with the requirement for a safe testing platform.
\item Can we show that controllers evolved through this improved optimisation are optimal for the mission/payload/UAV considered?
\end{enumerate}

 Results will guide the development of further research testbeds, and, as all current learning techniques are sensitive to the state-space the algorithm experiences, are broadly applicable to robot learning in general.

\subsection{Background}
To place our work in the broader context of relevant literature, we now briefly summarise research in evolving UAV control, and robot testbed development.

\subsubsection{Evolving UAV controllers}

EAs typically require a substantial number of evaluations to discover promising solutions due to the underlying mechanisms of iterative, population-based, stochastic search.  In addition, early generations are largely comprised of low-performance controllers.  Simulation, e.g.,~\citep{1688525,howard-snn-quad,Koppejan:2009:NRL} is therefore the preferred methodology when evolving UAV controllers, as (i) it cannot physically damage the UAV, (ii) the environment is fully controllable, and (iii) evaluations are parallellisable and generally run faster than realtime.

All simulations, no matter how complex, necessarily abstract reality to some degree.  As such, a continuing research focus is to cross the 'reality gap'~\citep{jakobi1995noise}, and replicate simulated performance on real robot.  Despite continuing efforts, including coevolutionary model learning~\citep{sussex23829}, specifically selecting for transferability~\citep{Koos:2010:CRG:1830483.1830505}, and manual post-evolution rule-tweaking~\citep{7412843}, the only way to guarantee performance in reality is to evolve in reality.

 Real-world attempts to evolve UAV controllers are few in number due to inherent difficulties of stochastic optimisation in hardware.  Control of a blimp is successfully evolved~\citep{wheels-to-wings} in a large, open space, but the slow dynamics of the blimp simplifies the control problem, as well as trivialising recovery from dangerous states.  Height and yaw control of a miniature helicopter~\citep{4983001} are evolved, although other degrees of freedom are neglected to prevent suboptimal controllers from potentially damaging the UAV.  Although there have been some attempts to perform exhaustive analysis of evolved solutions on real systems~\citep{4983001}, it is not a well-studied area of research.

\subsubsection{Robotic testbeds}

Real-world evolution implies the need for a robotic testbed, as defined in in Section~\ref{sec_intro}.  Testbeds provide a controllable, measurable evaluation environment.  They are chiefly used for field testing; ensuring that the system can handle real-world (i) sensor and actuator noise~\citep{how2008real}, (ii) software issues~\citep{nishiwaki2000design}, (iii) environmental conditions (e.g., underground~\citep{acar2001path}, space~\citep{samuele2010progress}), and (iv) mission requirements (e.g., multi-robot coordination~\citep{acar2001path}).  They focus on state estimation to evaluate performance, and the robots are typically manually controlled for missions with a duration from 5 minutes to a couple of hours.

Performing iterative optimisation in addition to evaluation imposes additional requirements on the testbed.  Such testbeds should consider (i) repeatable evaluations to ascertain fair, reliable optimisation scores, (ii) automated experimental management, to reduce the human requirement through multiple optimisation generations and remove the requirement to manually operate the robot under evaluation, and (iii) provision of power, again to reduce human intervention whilst making the optimisation tractable in time.

One of the earliest examples (1994) approximated a differential-drive ground robot with a gantry-mounted camera, thus removing issues with motor wear and power supply~\citep{harvey1994seeing}.  Technology advancements later allowed real robots to evolve on a testbed ---  examples include gait optimisation for legged robots~\citep{yosinski2011evolving,degrave2015transfer}.  More recent work includes multi-objective optimisation of legged robot behaviours as a fully closed-loop system~\citep{heijnen2017icra}, and a testbed that can alter its layout via a robot arm and vision/marker system~\citep{auto_increm_evo}, to evolve controllers for difficult environments by incrementally bootstrapping them in simpler ones.

Aside from~\citep{heijnen2017icra}, human intervention is required to, e.g., change batteries, reset the robot if it moves out of bounds, etc.  One notable example successfully evolves hardware UAV controllers~\citep{ghiglino2015online}, however state estimation requires an expensive infra-red tracking system, and frequent human intervention is required to change batteries.

Our testbed is fully described in~\citep{howard2017gecco,howard2017platform}, and is unique in that it can run back-to-back optimisations on a real UAV in a fair, repeatable manner, indefinitely, and without human intervention.   High-performance controllers are generated for real UAVs that are specific to the mission, payload, and hardware state of the UAV being optimised.  The most recent iteration of the testbed~\citep{howard2017gecco} optimises a controller for an arbitrary UAV (tested on two different quadrotors and a hexacopter) from scratch in 3-4 hours.  The testbed has also optimised control for a UAV carrying a heavy off-centre payload; something commercially-available self-tuning autopilots cannot do~\citep{howard2017platform}. 

As evaluations using the testbed occur in real-time, subsequent improvements focused on reducing the number of evaluations required.  Self-adaptive mutation rates, e.g., ~\citep{rechenberg}, are used as a form of experimental process optimisation, and adapt to the experimental specifics (UAV, payload, and environment) which significantly reduces the number of evaluations needed compared to algorithmically-determined static rate settings~\citep{howard2017platform}.  Rate restart strategies~\citep{howard2017gecco} make experiment times more predictable by reducing variance in the total number of evaluations required.

This research focuses on ensuring the feasibility of the controllers in real-world settings, including (i) generalisation analysis of evolved controllers, and (ii) methods that improve the ability of the controllers to function in the real world.  This work represents the final, critical, step before deployment for our UAV operations.

\section{Material and Methods}

\subsection{Testbed}
The testbed, which is exhaustively described in~\citep{howard2017platform},  comprises a solid floor which is covered with foam matting.  Before an experiment starts, the target UAV is set up in the testbed (Fig.~\ref{setup_new}).  Flipping (tilt angles $>60^o$), and excessive rotation ($\pm160^o$) of the UAV are prohibited by two nylon wires, which are attached to the UAV and the centre of the testbed's floor.  A camera, mounted centrally on a mesh-covered metal frame, provides position estimates.  Wind disturbances of $\approx$5m/s are provided by a fan, with a total traversal angle of $120^o$ and oscillation period of 10 seconds.  A 24V power tether permits continuous operation, and a serial cable connects to the host PC, which manages and monitors experiments using the real-time Extended State Machine (ESM) framework~\citep{merz2006control}.  Our testbed provides a comprehensive set of functions to enable closed-loop, automated UAV controller generation:

\begin{itemize}
\item {\bf Accurate state estimation} for control and evaluation, using a conventional camera (position) and AHRS (attitude), with no expensive marker-based tracking required.
\item {\bf Performance evaluation} of controllers in reality, in a confined space.
\item {\bf Robot-agnostic controller evolution} for any robot capable of hover, with no models or simulator required, automatically taking into account payload and hardware variability.
\item {\bf Continuous experimentation}, including 24/7 evolution, experimental monitoring and statistics recording or repeatable, fair tests.  The testbed can carry out back-to-back experiments for over a week without human intervention.
\item {\bf Health monitoring} of dangerous states, with UAV recovery.  Error detection algorithms identify and and re-run erroneous trials\footnote{Typical causes include, e.g., the tracking LED being obscured, or data link errors.}
\item {\bf Environmental interactions}, e.g., wind, can be included in the evaluations.\\
\end{itemize}

\subsection{Controllers \& State Estimation}
\begin{figure*}[h!]
\centering
 \subfloat{\includegraphics[width=12cm]{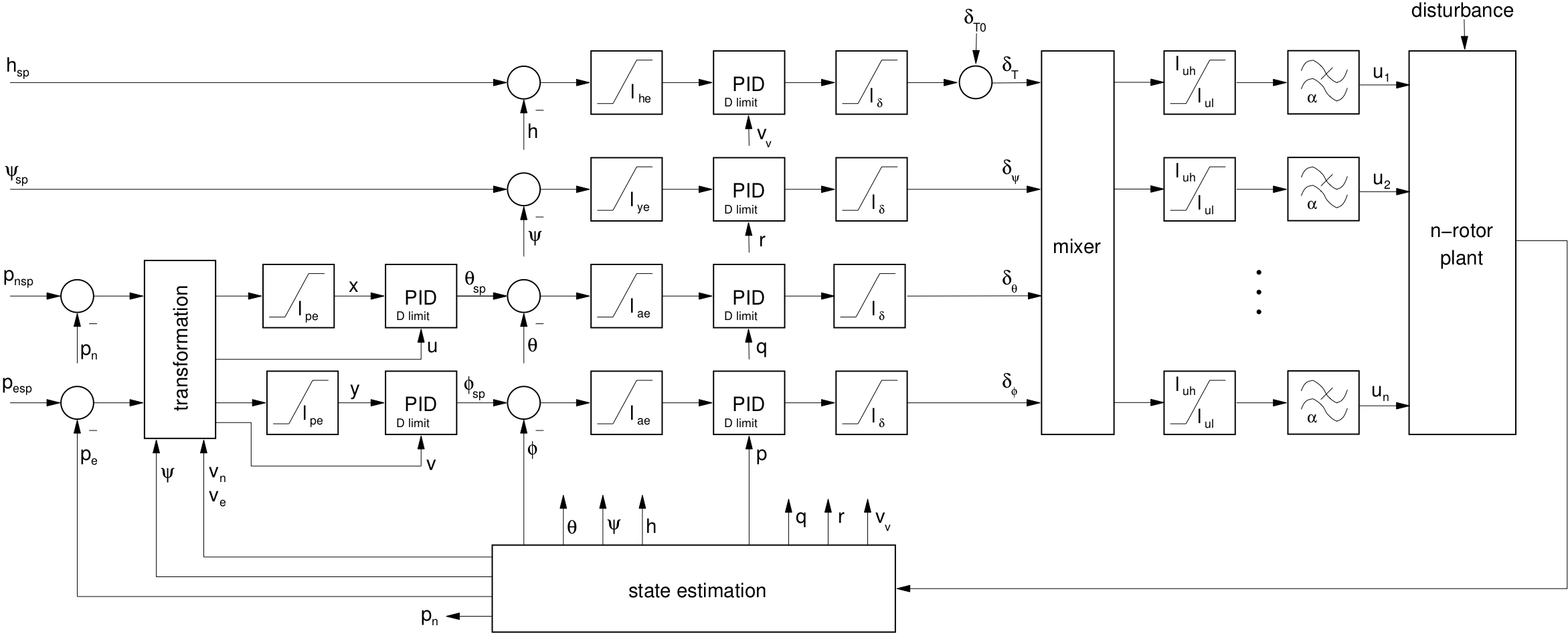} }
\caption[]{PID control structure, showing attitude and position loops.  Parameters $l_{he}$ $l_{ye}$ $l_{ae}$ denote error limits for height yaw and attitude respectively. $l_{ul}$ and $l_{um}$ are minimum and maximum motor commands, and $\delta_{\phi}$, $\delta_{\theta}$, $\delta_{\psi}$, and $\delta_{t}$ are command inputs to a mixer which outputs speed controller commands $u_1$ to $u_6$ (assuming a hexacopter). }
\label{PID-structure}
\end{figure*}

The goal is to optimise the PID controller of the UAV, which has a nested structure shown in Fig.~\ref{PID-structure}.  PID control aims to minimise the error of the UAV during flight, which we take as the deviation between the UAVs actual pose and desired pose.  Pose consists of 6 elements, or Degrees of Freedom (DoF);  roll $\phi$, pitch $\theta$, yaw $\psi$, height $h$, and lateral position in North $p_n$ and East $p_e$.   Three gains $K_p$, $K_i$, and $K_d$ control the error response per DoF.  Each controller is therefore parameterised by a {\em gain set} of 18 real-valued gains (6 DoFs, 3 gains per DoF).

At the start of an experiment, we bootstrap our population by repeatedly randomly initialising gain sets within allowed ranges (\ref{init-eq}). Here $l_\mathrm{cmd}$ is a generalised maximum possible command (Pulse Width Modulation, or PWM) for each parameter $V$: $l_\mathrm{cmd}$=300 for $\phi$, $\theta$, $\psi$, and $h$, and 20cm for $p_n$/$p_e$.  The gain set is accepted into the initial population if it allows the UAV to stay in the air for $0.2$s when attempting to hover at a height of 10cm with all over DoFs neutral.  When the population size reaches $N=20$, the first generation begins.

 \begin{equation}
 K_\mathrm{pV}=K_\mathrm{iV}=K_\mathrm{dV}=(0,\frac{l_\mathrm{cmd}}{l_\mathrm{er}}]
 \label{init-eq}
 \end{equation}

We evaluate each gain set by loading it into the PID structure.  To ensure a fair, repeatable test, the UAV is reset to a designated start position (in the centre of the floor area with $\psi$$\pm$60$^o$) between evaluations.  Once reset, the UAV attempts to fly a predefined waypoint set, where each waypoint defines a desired setting of position and yaw: ($\psi_\mathrm{sp}$, $P_\mathrm{nsp}$, $P_\mathrm{esp}$, $h_\mathrm{sp}$).  To successfully minimise error in the position waypoints, roll and pitch error must also be minimised; hence all 6 DoFs are optimised by these waypoints.  Note that the fan is not reset between trials; this prevents overspecialisation but provides different conditions to the controllers. To ensure a fair test we re-evaluate any successful controller 3 times in total to prove generality to wind conditions, and use an average fitness score (see Section\ref{sec:ea}).

As our controller responds to error, we must continually estimate the UAV's state to compare to our desired waypoint.  An onboard Inertial Measurement Unit (IMU, a Microstrain GX4-25) estimates roll $\phi$, pitch $\theta$, yaw $\psi$ (250Hz), and height $h$ (20Hz) when combined with an optical range finder.  An external camera estimates position in North $p_n$ and East $p_e$ at 60Hz.  Angular velocities $\dot{\phi}$, $\dot{\theta}$, $\dot{\psi}$ (250Hz) are derived from two consecutive Euler angles, and position velocities $\dot{p_n}$, $\dot{p_e}$, $\dot{h}$ (60Hz) calculated through a linear regression of five consecutive position estimates.  Height is processed through a complimentary filter; the Kalman filters integrated into the IMU were bypassed and manually re-implemented to give us full control and observability of the system.


The PID uses the state estimate to calculate errors $e$ in all DoFs.  Errors are taken as the difference between the waypoint value and the UAV's current value per DoF.  Each error is limited (10cm for $h$, 15$^o$ for attitude, 15cm for $p_n/p_e$) before being input to the PID.

The PID takes these error signals, and in response produces four outputs ($\delta_{\phi}$, $\delta_{\theta}$, $\delta_{\psi}$, and $\delta_{T}$), which represent commanded changes in roll, pitch, yaw, and thrust, scaled to fall in the range of possible motor PWMs ($l_{ul}$=1090 and $l_{uh}$=1950).   Error response follows (\ref{pid-eq}), where $o$ is the PID output, $t$ is the instantaneous time, and $\tau$ is the integration time step from 0 to $t$.  These outputs are passed to a linear mixer, which provides one control command $u$ per motor $m$, $u_1$ to $u_m$ at 250Hz.

\begin{equation}
o(t) = K_pe(t) + K_i \int_0^t e(\tau)d\tau + Kd\dfrac{d}{dt}e(t)
\label{pid-eq}
\end{equation}

\subsection{Evolutionary Algorithm}
\label{sec:ea}
During evaluation, a controller accumulates fitness (initially 0) at 250Hz by adding a per-Hz fitness measure $f_{cycle}$ (max. 10) to a running total $f$ (max. 150,000).  $f_{cycle}$ is a compound fitness function (\ref{eq:fitnessfunc}), with components measuring the error in pitch/roll ($f_{a}$), horizontal and vertical velocity ($f_{vh}$, $f_{vv}$), height ($f_{h}$), yaw ($f_{\psi}$), horizontal position ($f_{p}$), pitch/roll rates ($f_{\omega}$), and staying within controller limits ($f_l$).  Full calculations are shown in Appendix A.

\begin{equation}
f_{cycle} = f_{a} + f_{vh} +f_{vv} +f_{h}+f_{\psi}+f_{p} + f_{\omega} + f_{l}
\label{eq:fitnessfunc}
\end{equation}

Once each gain set has a fitness, we use self-adaptive Differential Evolution (DE)~\citep{de} to generate new gain sets.  DE is a state of the art optimiser of real-valued vectors, such as our PID gain sets~\citep{6217801, biswas2009design}, and has seen previous success in a robotic context~\citep{Moravec2014}; further justification for using DE can be found in~\citep{howard2017platform}.  

A donor vector {\bf v} is created for each gain set in the population {\bf p} following (\ref{de-eq}). $F$ is a differential weight, and {\bf r1}, {\bf r2}, and {\bf r3} are the gain sets of three unique randomly-selected individuals.

\begin{equation}
{\bf v} = {\bf r3} + F ({\bf r1} - {\bf r2})
\label{de-eq}
\end{equation}

A child controller {\bf c} is created for each parent by probabilistically replacing elements of {\bf p} with those of {\bf v}.  For each vector index $i$, $c_i$ = $v_i$ if $i==R$ or $rand$ $<$ {\em CR}, otherwise $c_i$ = $p_i$. $rand$ is a uniform-random number in range [0,1], and {\em CR} is the crossover rate.  $R$ is a vector index, selected randomly per {\bf c}, ensuring {\bf c} contains at least one element of {\bf v}.  Each child is evaluated and assigned a fitness $f$, and replaces its parent if it is fitter.  This concludes a generation.  Each subsequent generation involves creation of one child per parent, evaluating and assigning fitness, and creating the next generation by selecting parents and children based on fitness, as above.

Self-adaptation based on an Evolution Strategy~\citep{rechenberg} allows the testbed to tailor the learning process to the UAV, mission, and payload under consideration~\citep{howard2017platform}.  New population members random-uniformly initialise their {\em CR} and {\em F}, respecting bounds for {\em CR}=[0,1], and {\em F}=[0,2]. Each ${c}$ copies {\em CR} and {\em F} from its ${p}$, and modifies them following (\ref{sa_eq}), before using them to alter its controller.

\begin {equation}
\mu \leftarrow \mu * e^{N(0,1)}
\label{sa_eq}
\end {equation}

To prevent the learning process being stuck due to suboptimal rate settings, the rates of a given parent are uniform-randomly reinitialised if it does not create a fitter child within 5 generations\footnote{Selected to balance search stability and convergence times following a parameter sweep.}.  This combination of self-adaptation with restarts has been shown to promote high-fitness controllers, while significantly reducing the number of generations required, when compared to optimally-set static rates~\citep{howard2017gecco}.

As a stochastic optimser, we must be mindful of testing gain sets that are harmful to the UAV.  ESM monitors the UAVs behaviour throughout an evaluation, and safely terminates as required to preserve the UAV.  Terminated controllers are assigned their currently-accumulated fitness.  Termination occurs when;

\begin{itemize}
	\item angles $\phi>60^o$, $\theta/\psi>15^o$, 
	\item any angular rate$>250^o/sec$, 
	\item horizontal velocities $v_n$/$v_e>$ 50cm/s, 
	\item vertical velocity $v_h>$25cm/s, 
	\item total current draw $>20A$, 
	\item if the UAV does not take off within 5s of evaluation start, 
	\item or if the UAV pulls on the tether ($h>$18cm). \footnote{This latter criterion prevents the UAV from cheating by using the tether to 'balance' itself.}  
\end{itemize}

Any controller that successfully completes the entire waypoint set once is re-evaluated twice more, and assigned the average fitness of all three runs.  If the controller completes all three repeats, it is called a {\em success}.  The experiment ends when we have an entire population of successful controllers.

\subsection{Experimental}

Our testbed is designed to optimise UAV controllers for real-world flight.  Design decisions to date support this goal: optimising real UAVs, with real payloads, in strong, realistic wind disturbances.  However, stochastic optimisation of hardware UAV controllers for real-world conditions is challenging, as the amount of state space exploration must be balanced with the requirement for non-destructive controller evaluation.  Post-hoc analysis of some previously-generated controllers showed degraded performance when the UAV was removed from its tether and allowed to free-fly inside the test bed.  There are two main reasons for this.  Firstly, the state space available to the UAV may be too limiting, resulting in a lack of generalisation.  Secondly, controllers optimised in the 'ground' effect\footnote{when flying close the the floor, the ground deflects a propellers airflow, causing increased thrust nearer the ground for the same power input}~\citep{johnson2012helicopter}, may perform sub-optimally when outside of it.   A widely used ground-effect equation is shown in (\ref{ge-eq}), where $R$ is the radius of the rotor, $z$ is the vertical distance to the ground, $T$ is the thrust in ground effect, and $T_{\infty}$ is the thrust produced at the same power outside of the ground effect.

\begin{equation}
\displaystyle \frac{T}{T_{\infty}} = \frac{1}{1-\left(\frac{R}{4z}\right)^2}
\label{ge-eq}
\end{equation}

It follows that ground effect is negligible ($T \approx T_{\infty}$) when $z/R>$2\footnote{$>$4 in certain circumstances~\citep{powers2013influence}}, and that the influence of the ground effect diminishes rapidly the UAV gets further from the ground (\ref{ge-eq}).  The simplest way to mitigate both of these factors is to increase the amount of flying area available to the UAV, so TSE appears to be a simple and viable route towards real-world flyability. 

\begin{figure}[t]
\begin{center}
\includegraphics[width=9cm]{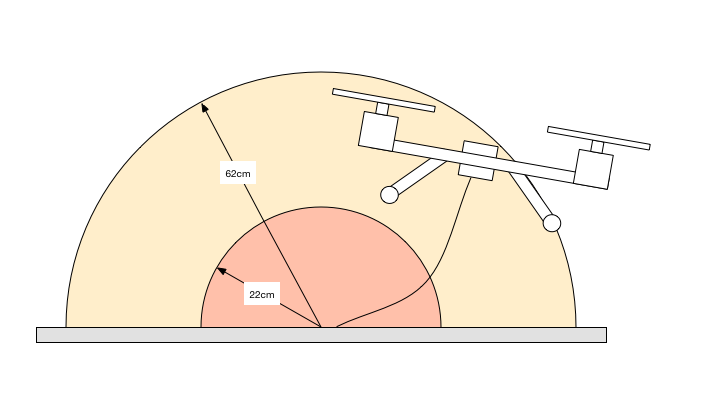}
\end{center}
\caption{Visualising the difference in flight envelope attainable in OSE (22cm, orange) and TSE (62cm, yellow).}
\label{hemispheres}
\end{figure}

In our first experiment, we compare the original approach of using a single physical tether throughout the experiment (One-Stage Evolution, OSE) with a new incremental method that partially evolves controllers with the original tether, before switching to a more permissive tether to complete the evolution in a larger state space (Two-Stage Evolution, TSE), as shown in Fig.~\ref{hemispheres}.

Controllers are optimised on a wind-affected hover scenario, with a total evaluation length of 60s. and waypoint transitions every 10s.  Transitions are rate-limited: $\psi=30^o/s$, N/E=0.1m/s, h=0.2m/s.  The waypoints are selected to excite all of the UAV's six DoFs by requiring controlled movements in each, and shown in Table~\ref{table:waypoints}.  

\begin{table}[t!]
\begin{center}
\caption{Waypoints for OSE (left in cell) and TSE (right in cell).  Position error is calculated from the centre of the UAV.}
\label{table:waypoints}
\begin{tabular}{lllll} 
 \noalign{\smallskip}\hline
t (s)   & N (m)       & E (m)   & h (m)   & $\psi$\\ 
  \noalign{\smallskip}\hline\noalign{\smallskip} 
 0    &0, 0     	&0, 0   		&0.2, 0.2 	&40, 40\\
10    &0.06, 0.15   &-0.06, -0.15	&0.2, 0.25 	&-5, -5\\
20    &-0.06, -0.15 &0.06, 0.15 	&0.2, 0.4 	&40, 40\\
30    &0.06, 0.15   &0.06, 0.15 	&0.2, 0.25  &85, 85\\
40    &0.06, 0.15   &0.06, 0.15 	&0.2, 0.4 	&40, 40\\
50    &0, 0     	&0, 0   		&0.2, 0.25  &40, 40\\
60    &stop     	&stop   		&stop   	&stop\\
 \noalign{\smallskip}\hline
\end{tabular}
\end{center}
\end{table}

The OSE tether is set to 22cm, the maximum length that prevents the UAV from flipping.  The TSE tether is set to 62cm, the maximum length that prevents the UAV from contacting the fan, to more closely represent free flight.  The experiments proceed as follows:

\begin{itemize}
\item OSE controllers fly the OSE waypoint set until all controllers are successful.
\item TSE controllers initially fly the OSE waypoint set until the population contains its first {\em successful} controller.  A new population ($N$=20) is created from copies of this controller, with each gain subject to uniform noise of $\pm$25\% of the total range of that gain when copied.  The UAV is transferred to a longer tether (Fig.~\ref{hemispheres}), and experimentation continues on TSE waypoints until all controllers are successful.  
\end{itemize}

Ten experimental repeats of OSE and TSE are run, with statistical significance assessed with a Mann-Whitney U-test (which does not require normally-distributed samples).  OSE and TSE repeats both have a population size of 20 controllers, and are run until the entire population is filled with successful controllers.

\section{Results and Discussion}

\begin{figure*}[t]
\begin{center}
\includegraphics[width=8cm]{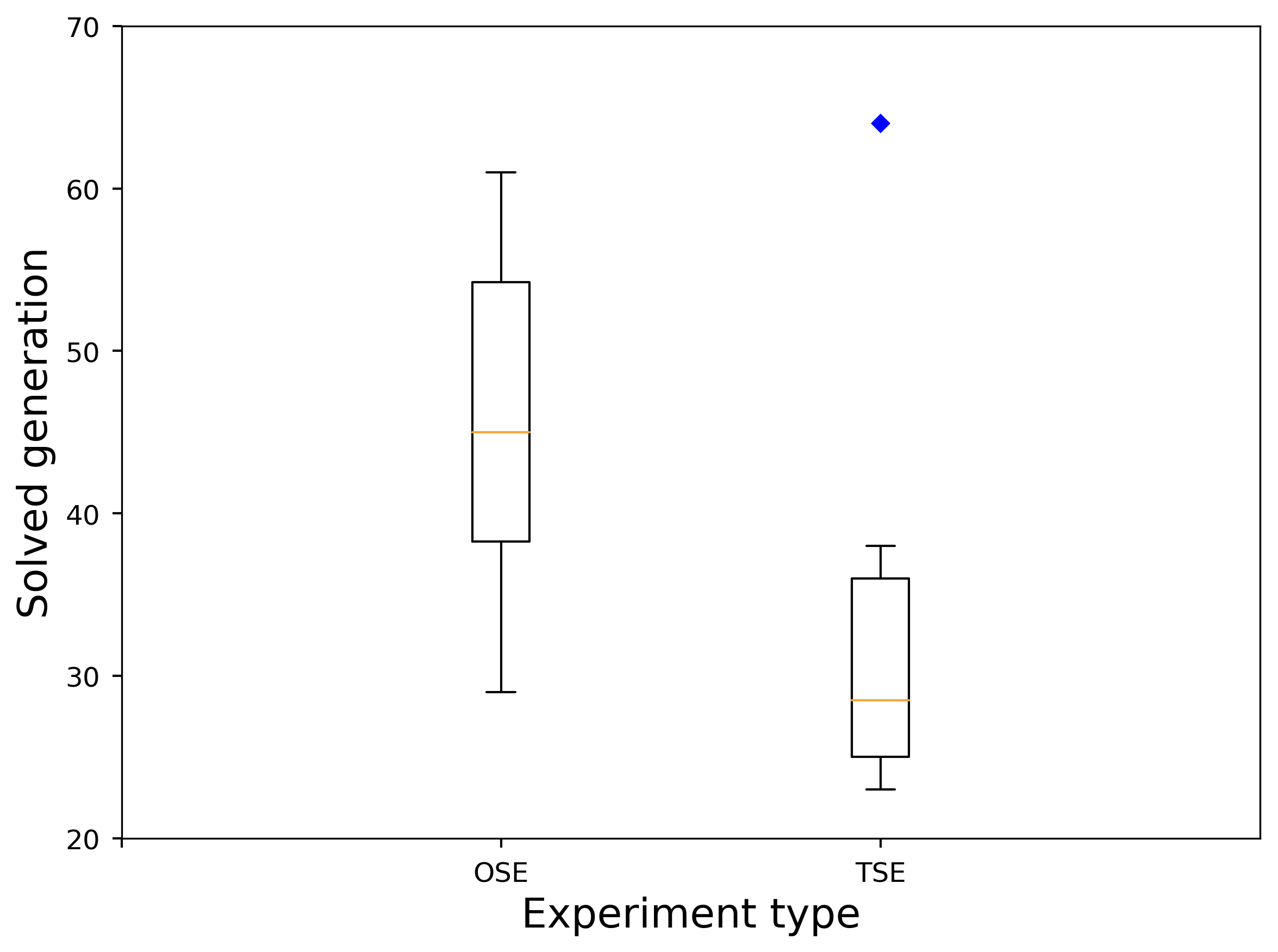}
\end{center}
\caption{A box plot comparing convergence performance for OSE and TSE.}
\label{graphxx}
\end{figure*}

\begin{figure*}[t]
\begin{center}
\subfloat[] {\includegraphics[width=7cm]{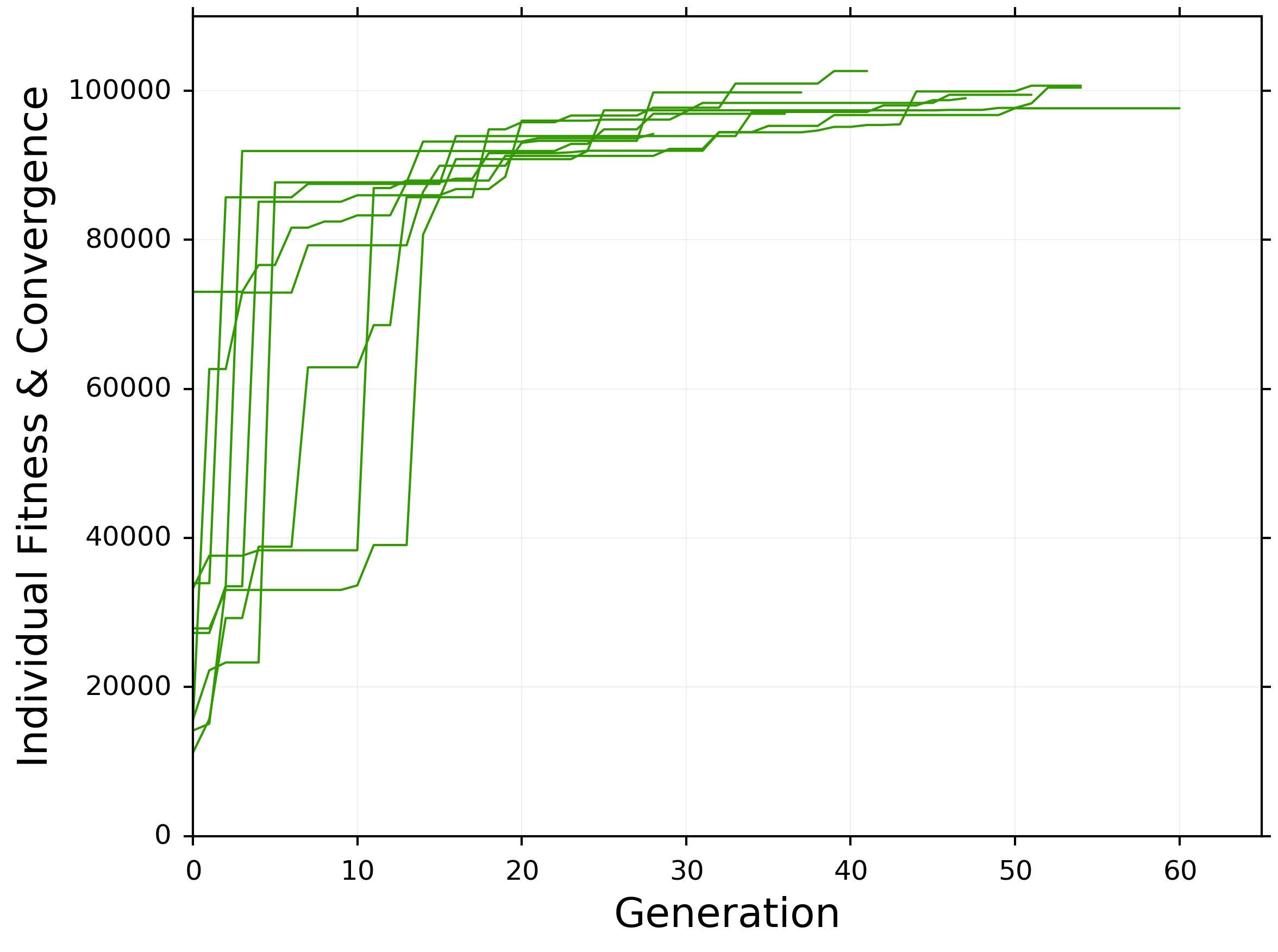}}
\subfloat[] {\includegraphics[width=7cm]{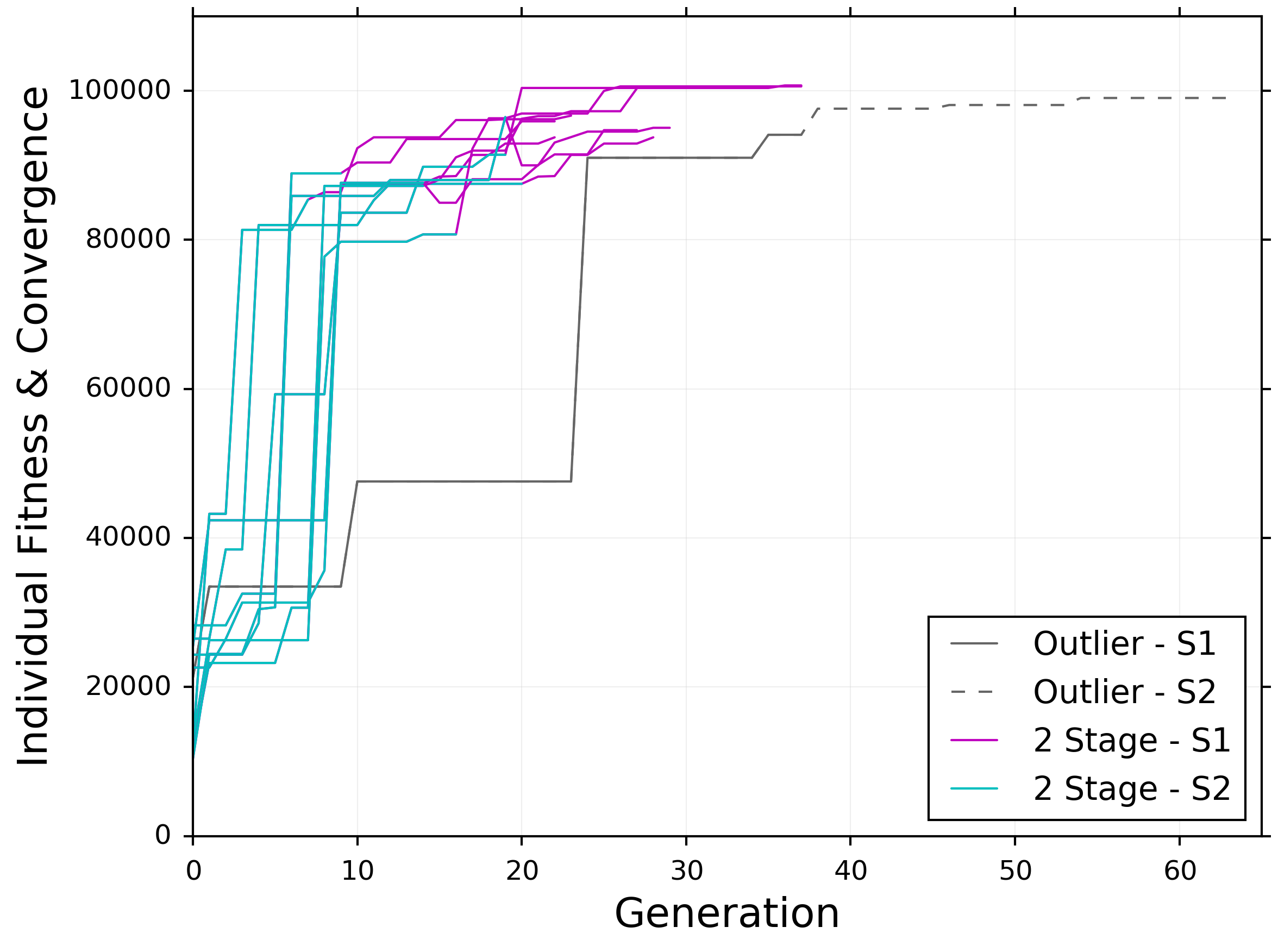}}
\end{center}
\caption{Showing individual fitness progressions for (a) OSE and (b) TSE in experiment 1.  Aside from a single outlier (grey line) in (b), OSE experiments typically take much longer to converge than their TSE counterparts.}
\label{graph3x}
\end{figure*}

As we operate in hardware and are limited to real-time evaluations, reducing the number of evaluations required to optimise our controllers is particularly interesting to us (see, e.g.,~\citep{howard2017gecco}).  Here, we see that TSE lowers the mean convergence generation from 45.6 (OSE) to 32.5 (Fig.~\ref{graphxx}, which is statistically significant (p$<$0.05) and indicates that TSE is a viable technique for reducing the number of generations required to perform an experiment.  Although generally more closely distributed around the mean, standard deviation for TSE convergence is higher than OSE (12.32 vs. 10.13) due to a single outlier at generation 64.  Visualising individual experimental fitness progressions reinforces this.  Although TSE typically converges faster than OSE (Fig.~\ref{graph3x}(a) vs Fig.~\ref{graph3x}(b)), the outlier in Fig.~\ref{graph3x}(b) (a grey line) lags far behind the other repeasts.  Despite this, TSE is clearly preferable in terms of convergence.

Both OSE and TSE produce high fitness controllers (Fig.~\ref{graph3x}(a) and (b)).  Final fitness values are quantatively similar, but cannot be directly compared as consecutive TSE waypoints are typically further from each other than OSE waypoints.   Mean fitness for OSE and TSE controllers was 99234 vs. 97184, 93090 vs. 90880, and 82786 vs. 82735, for best, average, and worst fitness respectively.

As a more representative test of the generalisation ability of the controllers,  we run the highest fitness controller per repeat on an unseen waypoint set.  Each controller in this set would be selected for any real-world deployment, based on their performance in the first experiment.

This waypoint set has the following transitions, with one transition per 10s as before.  The four array elements respectively represent position in N(m) and E(m), height h(m), and yaw $\psi$: ( [0.0, 0.0, 0.4, -10], [-0.25, 0.25, 0.2, 45], [0.25, -0.25, 0.4, 85], [0.25, 0.25, 0.4, 85], [0.0, 0.0, 0.3, -10], [-0.25, -0.25, 0.4, 45]).  The UAVs are placed on the longer stage 2 TSE tether.  Each controller is evaluated 20 times to generate reliable results.  The UAV flies away from significant ground effect.



\begin{figure*}[t]
\begin{center}
\centerline{
\subfloat[] {\includegraphics[width=7cm]{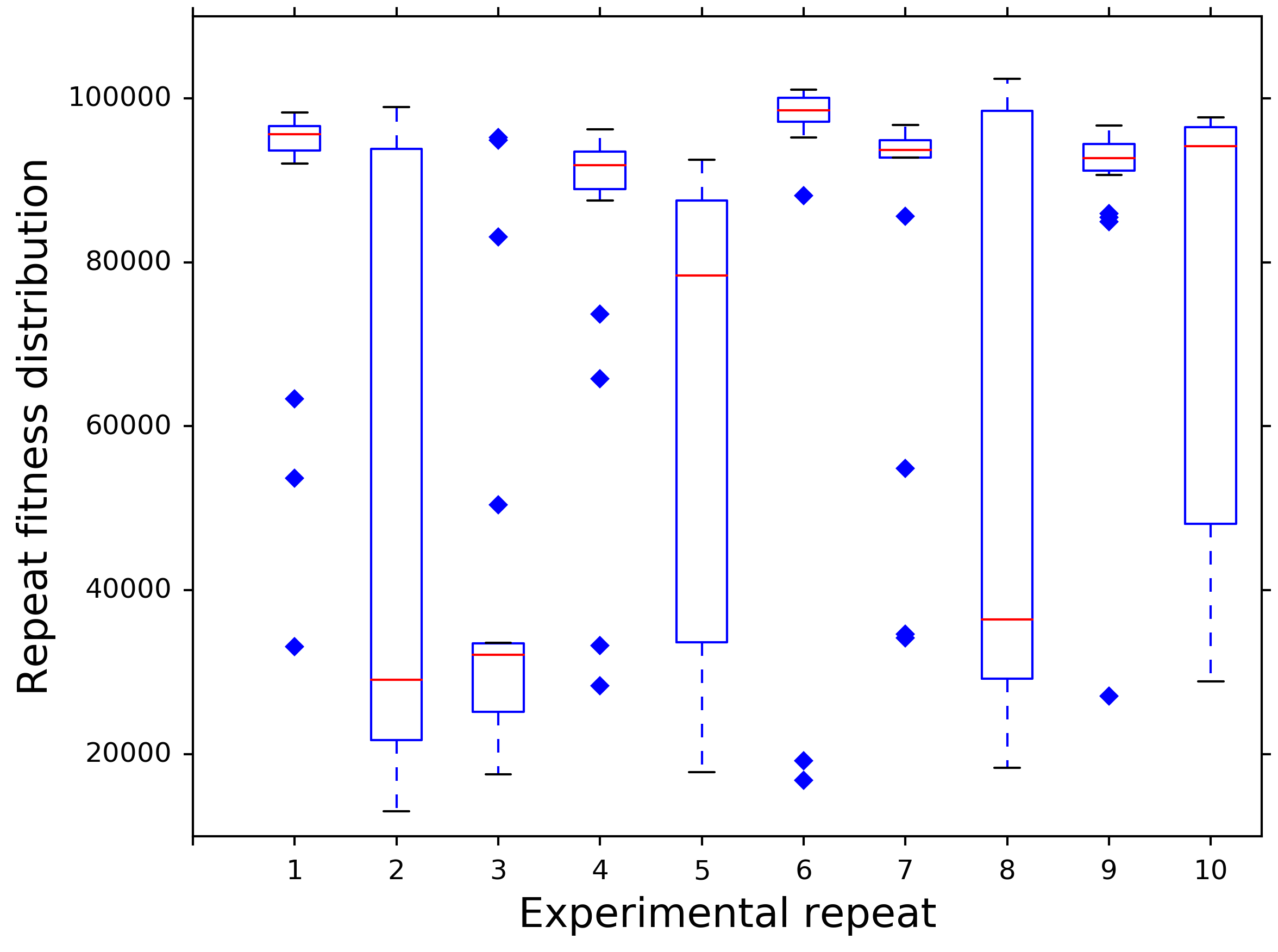}}
\subfloat[] {\includegraphics[width=7cm]{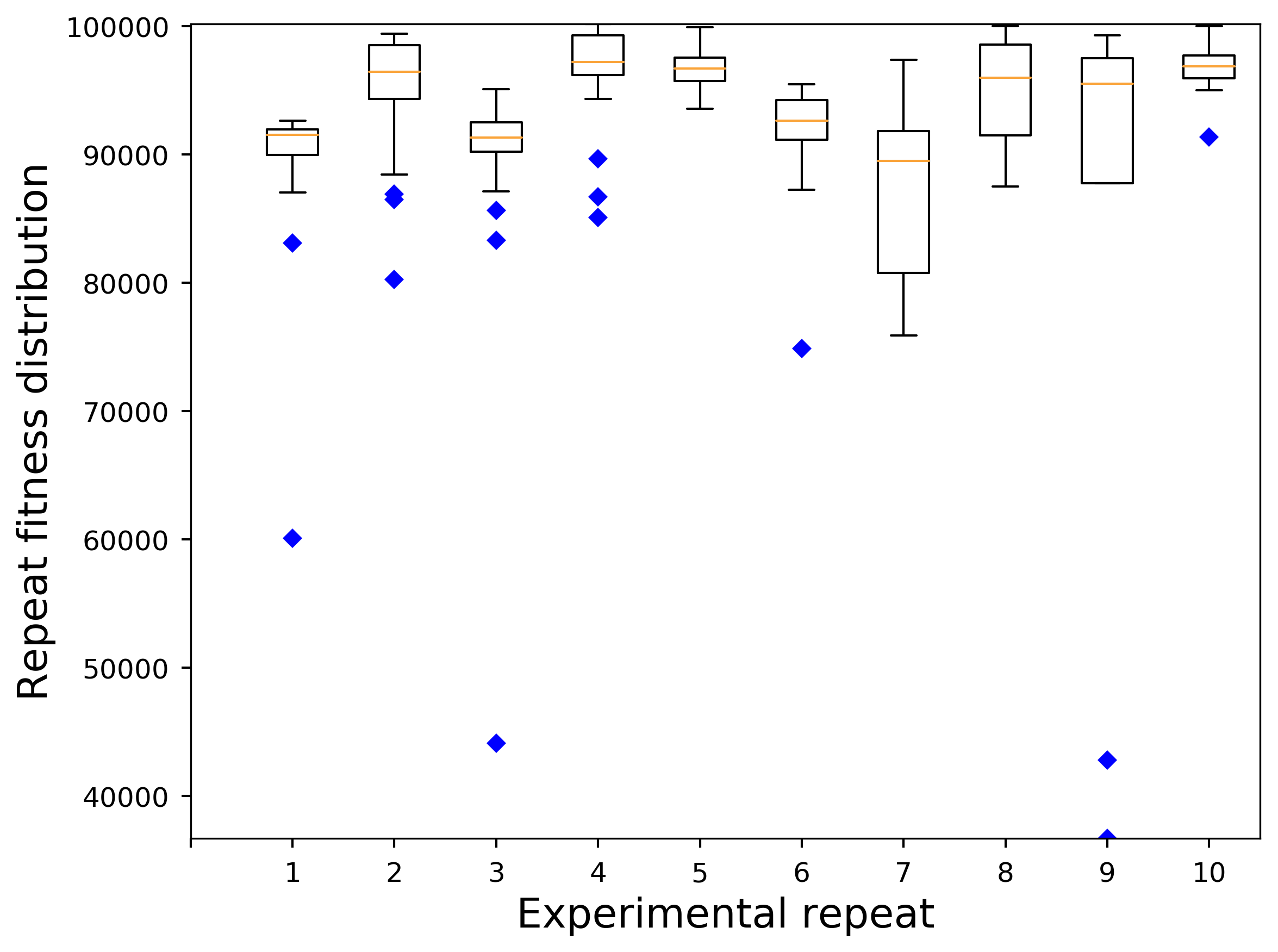}}
}
\end{center}
\caption{Showing the mean fitness for 20 repeats of the best (a) OSE and (b) TSE controllers per experimental run, when evaluated on the unseen waypoint set.}
\label{graph5}
\end{figure*}

Results are shown in Fig.~\ref{graph5}.  Notably, half of the OSE controllers (those from repeats 2, 3, 5, 8, and 10, Fig.~\ref{graph5}(a)) are seen to struggle when flying the TSE waypoints on a longer tether.  This suggests that the OSE setup does not properly expose the required state space to the controller, and as such some OSE controllers struggle when flying the new waypoint set.  All TSE controllers (Fig.~\ref{graph5}(b)) can handle the TSE waypoint set, displaying a consistently higher fitness than OSE controllers.

 It was observed that many OSE controllers from the aforementioned repeats oscillate, which causes premature fight termination and the resulting low fitness values.  This is likely due to the restricted state space available during OSE experiments, and indicates that OSE is not the best way to evolve controllers for real-world scenarios.  TSE is shown to be more successful in producing such controllers by being able to provide a more expansive state space during evolution.  However, it should be noted that both OSE and TSE are capable of generating real-world flying controllers (OSE repeats 1, 4, 6, 7, and 9 in Fig.~\ref{graph5}(a)).  

The best fitness score overall for OSE was 102363.1, and for TSE 100188.1.  Averaged across the entire experimentation on the new waypoint set, OSE fitness was 69810.6, and TSE fitness was 91839.7.  Averaging each controller's repeats individually, this is statistically significant.

\begin{figure*}[t]
\begin{center}
\centerline{
\subfloat[] {\includegraphics[width=7cm]{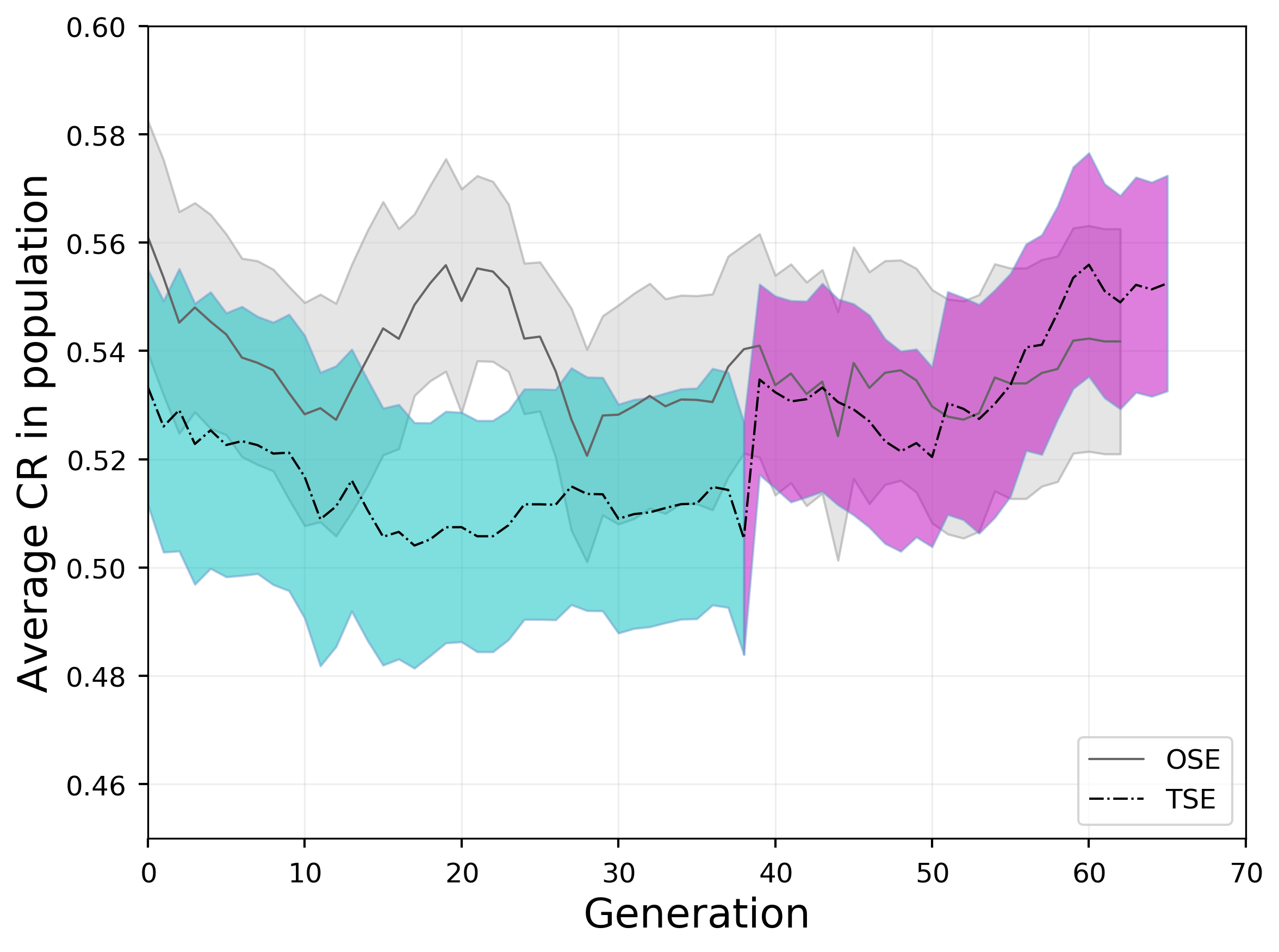}}
\subfloat[] {\includegraphics[width=7cm]{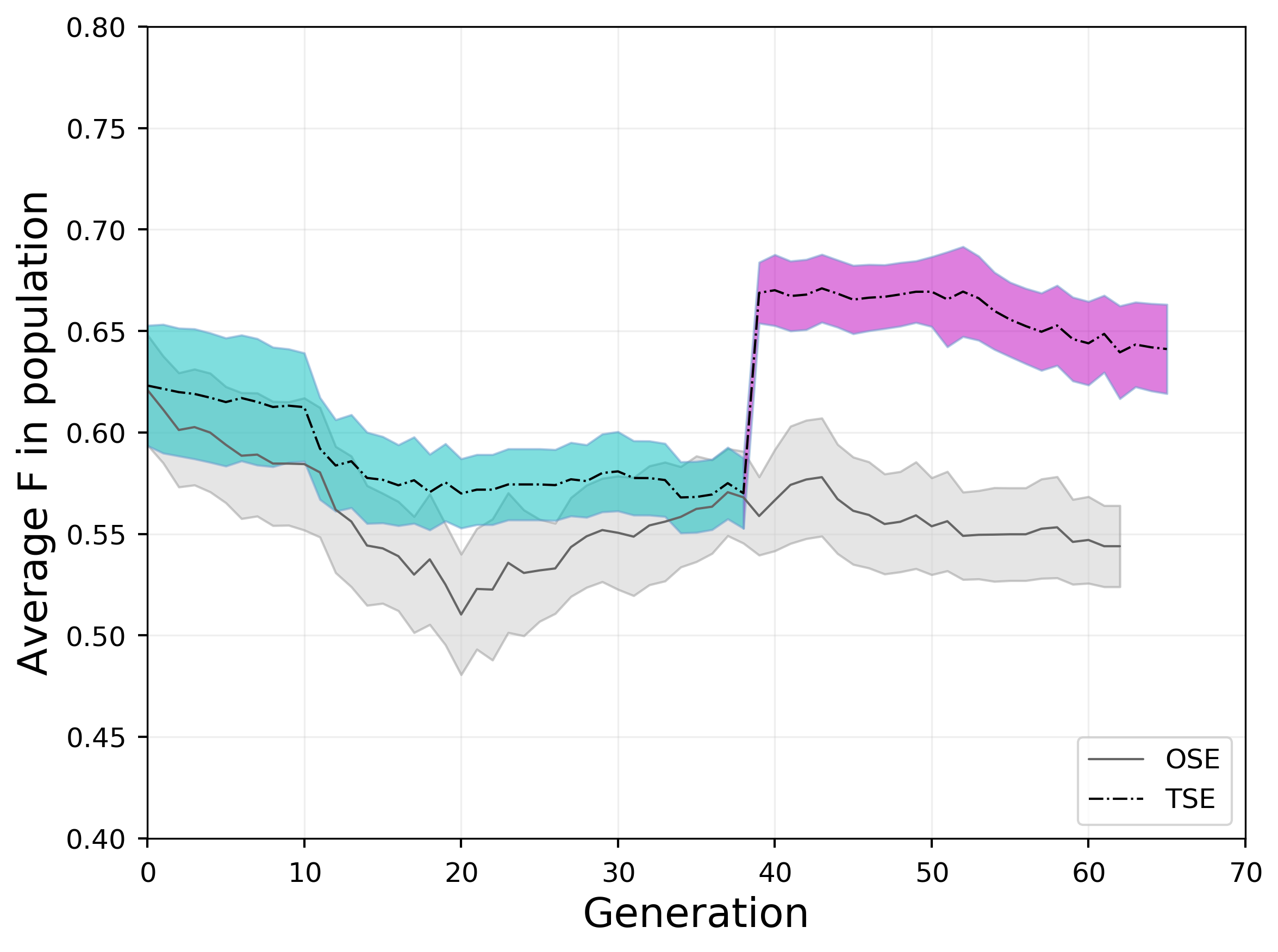}}
}
\end{center}
\caption{Self-adaptive rates $CR$ and $F$ in experiment 1 for OSE and TSE.}
\label{fig:adaptation}
\end{figure*}

Next, we assess the affect of the TSE experimental procedure on our self-adaptive mutation strategy.  The mean crossover rate $CR$ varies between 0.482 and 0.553 for the first stage of TSE, raising to a maximum of 0.589 for the second stage (Fig.~\ref{fig:adaptation}(a)).  Context-sensitive adaptation of the learning rates is seen between stage 1 and 2 in TSE, in particular the rise in $CR$ is a response to the reseeding of controller parameters, which subsequently requires more parent-child variance.  Self-adaptive rates also vary between OSE and TSE, showing again that rates can adapt to experimental setup as required.

$F$ has more effect on the evolutionary process than $CR$; as such the setting of $F$ between stage 1 and 2 in TSE is more pronounced than the corresponding change observed between stages for $CR$, which agrees with the literature of rate setting in DE\citep{de}, and supports previous results for our testbed~\citep{howard2017platform}.  $F$ self-adapts from 0.57 to 0.672 within 2 generations of stage 2 commencing, which induces more variance in the search process, either (i) in response to the increased search space, or (ii) to combat the increased parameter convergence in the stage 2 population.  This feature is key to TSE achieving more expedient convergence than OSE. Stage 2 $F$ rates are statistically different from both stage 1 and OSE $F$ rates.

\subsection{Controller Optimality and Sensitivity}

Next, we incrementally step through a range of feasible gain settings to show the optimality of our discovered controllers.  We focus on height control as it is most critical to the UAVs ability to accumulate fitness.  Mapping the effects of changing gains on controller fitness allows us to (i) confirm that our testbed produces the highest fitness controllers possible, and (ii) estimate how brittle those controllers are.  Note that without the testbed, this mapping would be extremely taxing to achieve.

The highest fitness controller from TSE is used;  we take the three height gains for that controller (P, I, and D), and for all perturbations, hold one gain static whilst systematically varying the other two gains within the following ranges:

\begin{itemize}
\item P gains step from -0.1 to -1, in increments of 0.1.
\item I gains step from -0.15 to -1.5, in increments of 0.15.
\item D gains step from -0.05 to -0.5, in increments of 0.05.
\end{itemize}

\begin{figure}[]
\begin{center}

\subfloat[] {\includegraphics[width=8cm]{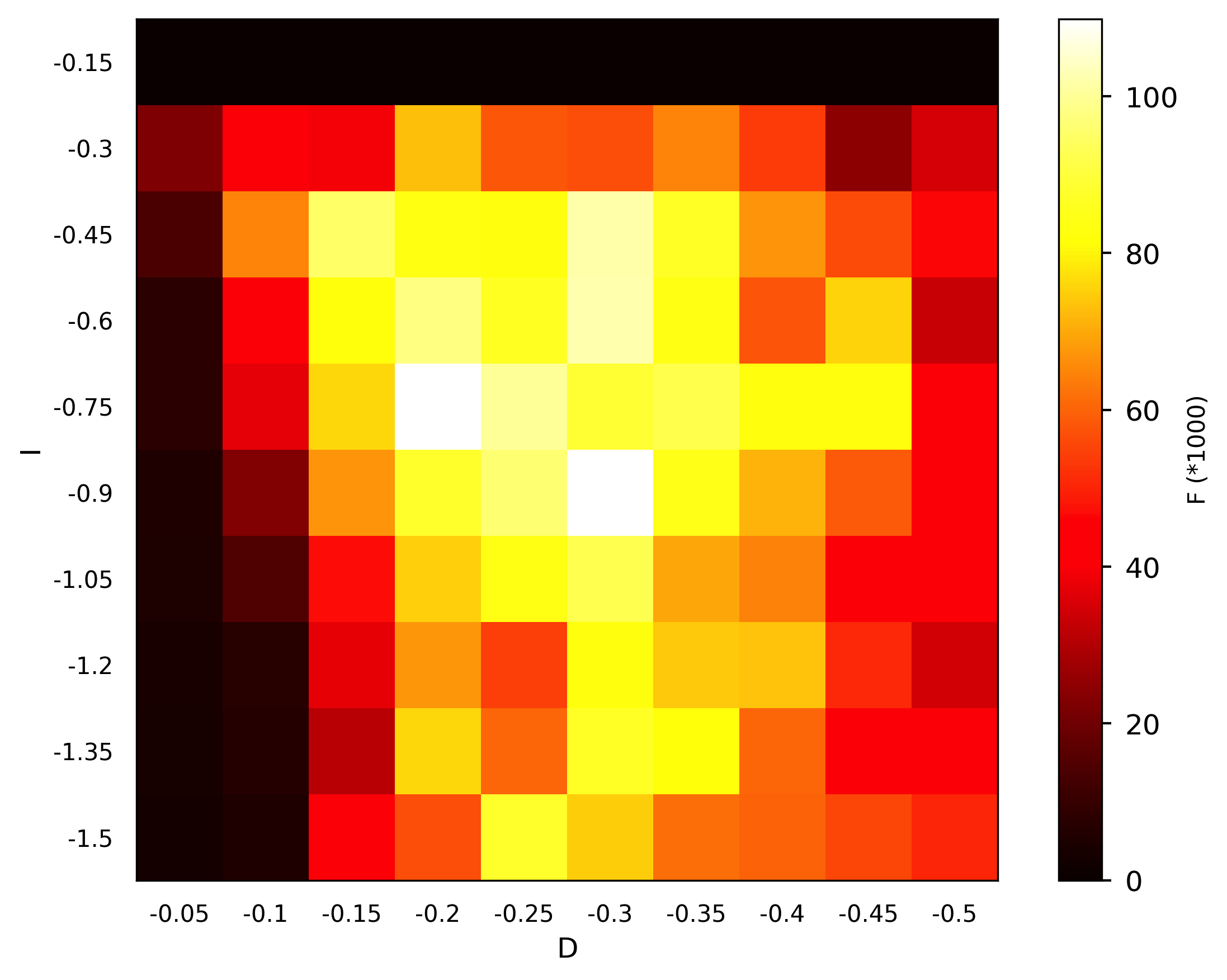}}\\
\subfloat[] {\includegraphics[width=7cm]{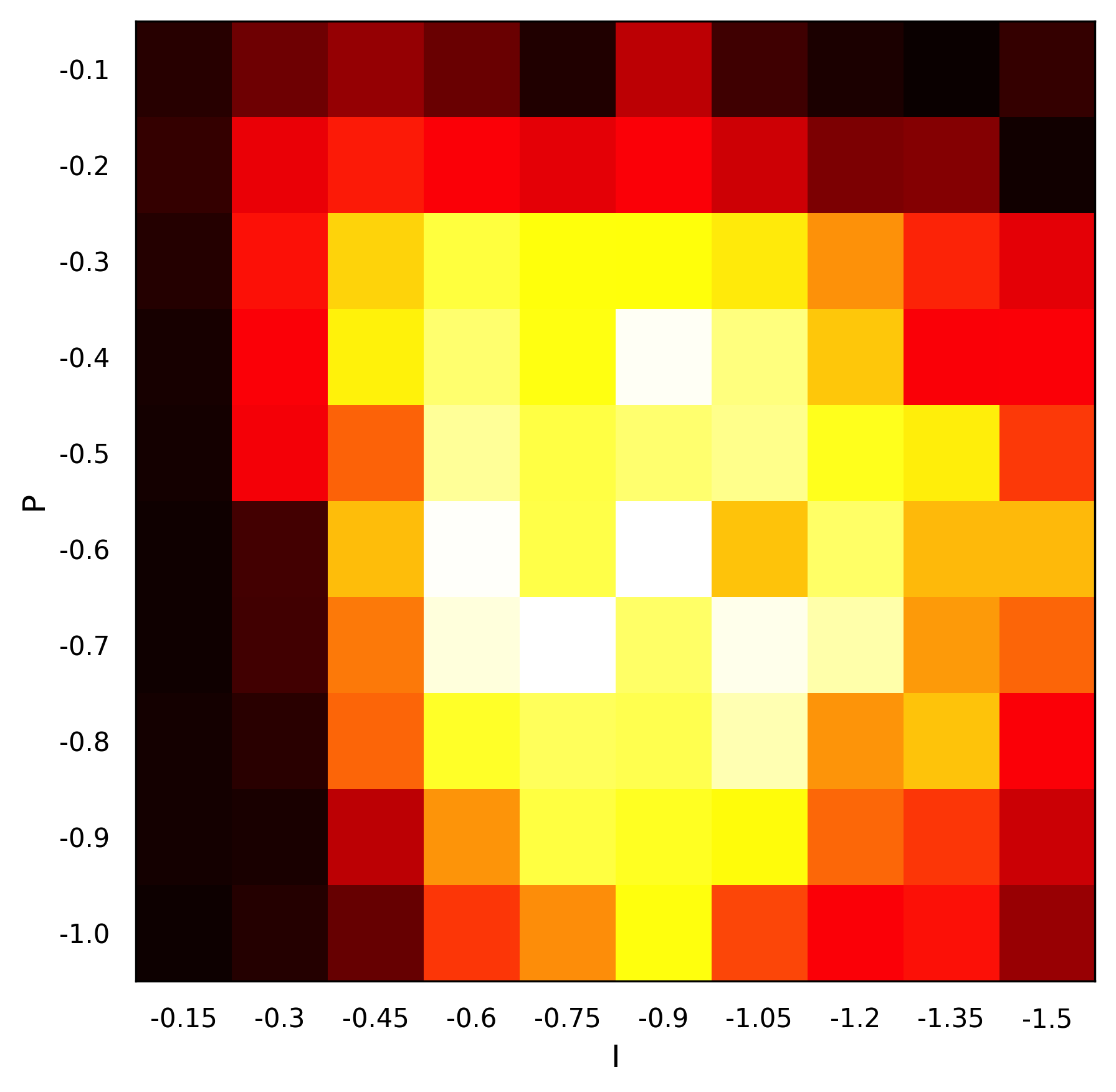}}
\subfloat[] {\includegraphics[width=7cm]{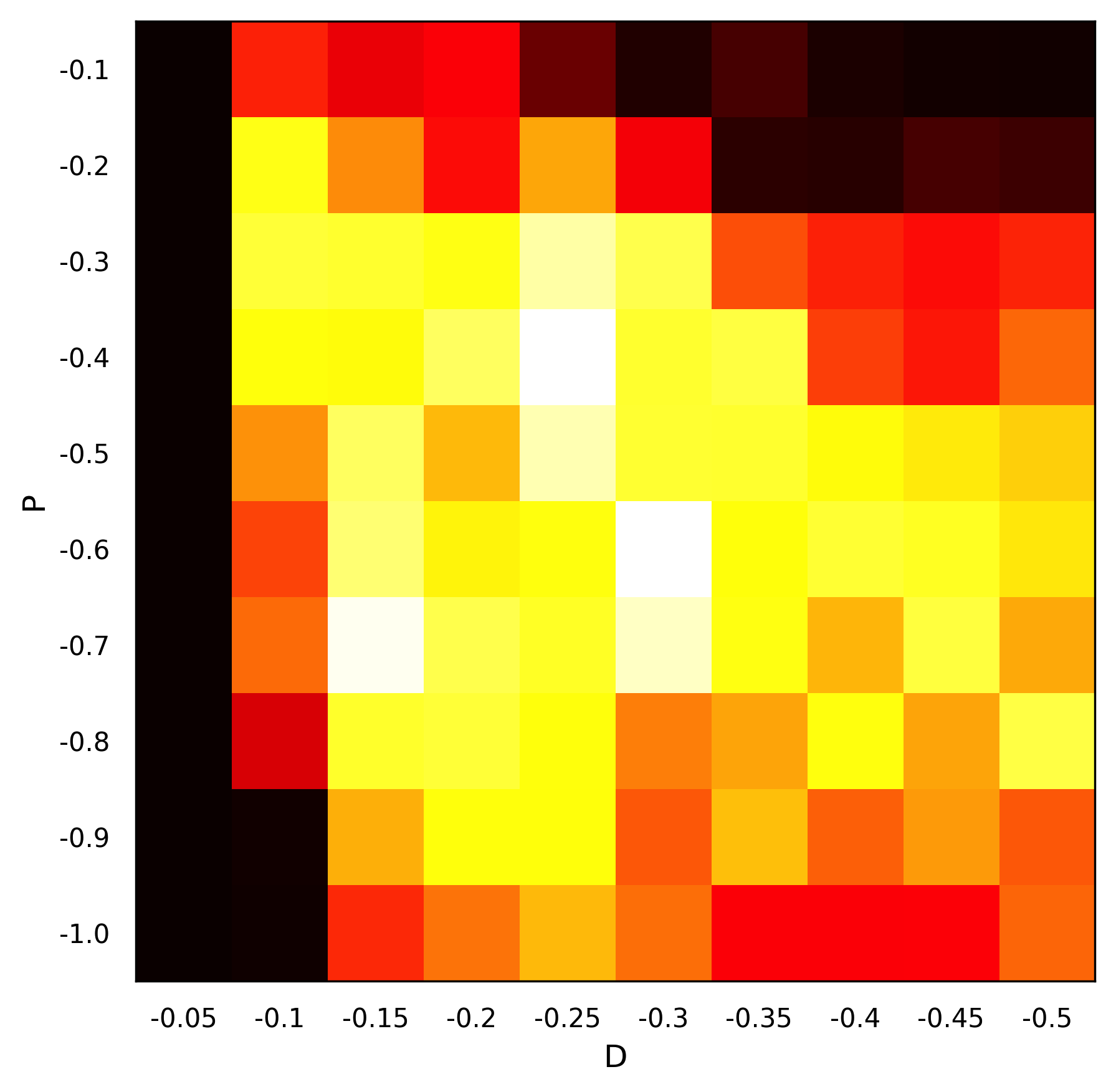}}

\end{center}
\caption{(a)-(c) fitness heat maps obtained from varying gain combinations I\&D, P\&I, and P\&D respectively, for height control.}
\label{mapzfit}
\end{figure}

This gives us ten possible settings per gain.  Ranges were chosen to fall centrally around the corresponding value from the best evolved controller, whilst covering a wide range of possible rate settings.  Each gain combination is run 20 times on the final waypoint set using the TSE tether, and fitness values are recorded and averaged.  The fitness landscape produced by this exhaustive gain search (Fig.~\ref{mapzfit}) shows performance degradation when gains diverge from their optimal values.

Notably, the maximum fitness values discovered through this search reach within 3$\%$ of the best evolved controller's fitness; in other words, evolution consistently finds the best controller settings.  Owing to the repeatability offered by the testbed, we can confirm that our best evolved height controller lies within the discretised global maximum discovered through exhaustive search (although in significantly less time, with fewer evaluations required).

Mapping the search space in this way also allows us to easily see infeasible solution subspaces, e.g. low I (-0.15) where the UAV is unable to take off before being `timed out' after 5s of inactivity, or low D (-0.05) gains causing trial-ending overshoots; which is an important feedback for future experimental design in terms of being able to ignore certain solution subspaces {\em a priori}.  The complexity of the problem (through e.g., sensory state estimation, noisy fitness evaluations, and other real-world artefacts) is evidenced through the appearance of multiple disconnected local maxima, shown in Fig.~\ref{mapzfit}(b) and (c).

\subsection{Conclusions} 

In this paper we investigated a method to increase the real-world flyability of our UAV controllers by increasing the amount of state space, and reducing the ground effect.  Wße compared our standard method of restrictive tethering (OSE) to a new method of partially evolving the controllers, and subsequently transferring to a longer tether (TSE) for extra exploration.  

Statistical comparisons between OSE and TSE shows that TSE produces {\em more general} controllers than OSE.  As such, TSE will be the new {\em de facto} technique as we  continue our experimentation on the testbed.  We note that this technique can potentially be extended with various tether lengths, where the incremental learning we demonstrate here allows for the safe transition to progressively larger and larger state spaces.

Finally, we used the testbed to perform an exhaustive gain analysis, and showed that our testbed successfully finds the global optimum gain settings for the scenario we investigated.  

These findings motivate further development towards our final goal quickly optimising high-performance controllers for arbitrarily-configured UAVs, that are optimised to mission, payload, and morphology, and are guaranteed to work in the real world.  In particular, we see an opportunity for our testbed in optimising the control and behaviour UAVs with arbitrary morphologies.  Such UAVs could be designed through evolution, and come in a range of unconventional physical configurations that may not be easily modellable.  Optimisation in our testbed guarantees real-world performance and does not require a model, and as such will be a prominent technique for allowing such UAVs to reach their potential.




\section {Funding Sources}
This research received funding from the CSIRO Office of the Chief Executive for the Postdoctoral position in Evolutionary Aerial Robotics.

\section{Conflicts of Interest}
Gerard David Howard declares no conflict of interest.  Alberto Elfes declares no conflict of interest.

\section*{Appendix A: Fitness Function}

\noindent
\begin{math}\displaystyle
f_\mathrm{cycle}=f_{\mathrm{p}}+f_{\mathrm{h}}+f_\psi+
f_{\mathrm{a}}+f_{\mathrm{vh}}+f_{\mathrm{vv}}+f_{\mathrm{\omega}}+f_{\mathrm{l}}
\\[1em]
f_\mathrm{p}=f\left(\sqrt{(p_\mathrm{nsp}-p_\mathrm{n})^2+(p_\mathrm{esp}-p_\mathrm{e})^2},l_\mathrm{p},l_\mathrm{pc}\right)
\\
f_\mathrm{h}=f(|h_\mathrm{sp}-h|,l_\mathrm{h},l_\mathrm{hc})
\\
f_\mathrm{\psi}=f\left(|\mathrm{wrap}(\psi_\mathrm{sp}-\psi)|,l_\psi,l_{\psi\mathrm{c}}\right)
\\
f_\mathrm{a}=f(|\phi_\mathrm{sp}-\phi|,l_\mathrm{a},l_\mathrm{ac})+
f(|\theta_\mathrm{sp}-\theta|,l_\mathrm{a},l_\mathrm{ac})
\\
f_\mathrm{vh}=f\left(\sqrt{v_\mathrm{n}^2+v_\mathrm{e}^2},l_\mathrm{vh},l_\mathrm{vhc}\right)
\\
f_\mathrm{vv}=f(|v_\mathrm{v}|,l_\mathrm{vv},l_\mathrm{vvc})
\\
f_\mathrm{\omega}=f(|p|,l_\omega,l_{\omega\mathrm{c}})+f(|q|,l_\omega,l_{\omega\mathrm{c}})
\\
f_\mathrm{l}=1-\frac{\mbox{number of controllers exceeding $l_\delta$ limits}}{4}
\\[1em]
f(e,l,l_\mathrm{c})=
\begin{cases} 
\mathrm{max}\{\frac{l-e}{4(l-l_\mathrm{c})},0\}, & \mbox{if }
e>l_\mathrm{c} \\
\frac{3(l_\mathrm{c}-e)}{4 l_\mathrm{c}}+\frac{1}{4}, & \mbox{otherwise}
\end{cases} \\[1em]
\mathrm{wrap}(\gamma)=\mathrm{atan2}(\sin(\gamma),\cos(\gamma))
\end{math}

\subsubsection*{Symbol Definitions - Fitness Function}
\noindent
\begin{math}\displaystyle
f_\mathrm{cycle}: \mbox{fitness for one control cycle}\\
f_\mathrm{p}/f_\mathrm{h}: \mbox{fitness for position/height tracking}\\
f_\psi: \mbox{fitness for yaw angle tracking}\\
f_\mathrm{a}: \mbox{fitness for pitch and roll angle tracking}\\
f_\mathrm{vh}/f_\mathrm{vv}: \mbox{fitness for low horizontal/vertical
velocity}\\
f_\mathrm{\omega}: \mbox{fitness for low pitch and roll rates} \\
f_\mathrm{l}: \mbox{fitness for staying within controller limits}
\\[1em]
p_\mathrm{n}/p_\mathrm{e}: \mbox{north/east position}\\
p_\mathrm{nsp}/p_\mathrm{esp}: \mbox{north/east position setpoint}\\
h: \mbox{height above ground}\\
h_\mathrm{sp}: \mbox{height setpoint}\\
\phi/\theta/\psi: \mbox{roll/pitch/yaw angle}\\
\phi_\mathrm{sp}/\theta_\mathrm{sp}/\psi_\mathrm{sp}: \mbox{roll/pitch/yaw setpoint}\\
v_\mathrm{n}/v_\mathrm{e}/v_\mathrm{v}: \mbox{north/east/vertical velocity}\\
p/q/r: \mbox{roll/pitch/yaw rate}
\end{math}

\begin{math}\displaystyle
l_\mathrm{p}: \mbox{position error range (0.2m)}\\
l_\mathrm{pc}: \mbox{core position error range (0.07m)}\\
l_\mathrm{h}: \mbox{height error range (0.2m)}\\
l_\mathrm{hc}: \mbox{core height error range (0.03m)}\\
l_{\psi}: \mbox{yaw error range (30$^o$)}\\
l_{\psi c}: \mbox{core yaw error range (10$^o$)}\\
l_\mathrm{a}: \mbox{attitude angle error range (15$^o$)}\\
l_\mathrm{ac}: \mbox{core attitude angle error range (3$^o$)}\\
l_\mathrm{vh}: \mbox{horizontal velocity error range (1 m/s)}\\
l_\mathrm{vhc}: \mbox{core horizontal velocity error range (0.2m/s)}\\
l_\mathrm{vv}: \mbox{vertical velocity error range (1 m/s)}\\
l_\mathrm{vvc}: \mbox{core vertical velocity error range (0.1 m/s)}\\
l_\mathrm{\omega}: \mbox{pitch and roll rate error range (250$^o$/s)}\\
l_\mathrm{\omega c}: \mbox{core pitch and roll rate error range (30$^o$/s)}\\
\end{math}

\bibliographystyle{spbasic}
\bibliography{evin}

\end{document}